\documentclass[10pt,twocolumn,letterpaper]{article}

\usepackage{cvpr}              

\usepackage[dvipsnames]{xcolor}

\definecolor{cvprblue}{rgb}{0.21,0.49,0.74}
\usepackage[pagebackref,breaklinks,colorlinks,citecolor=cvprblue]{hyperref}

\def\paperID{12031} 
\def\confName{CVPR}
\def\confYear{2024}

\usepackage[utf8]{inputenc} 
\usepackage[T1]{fontenc}    
\usepackage{hyperref}       
\usepackage{url}            
\usepackage{booktabs}       
\usepackage{amsfonts}       
\usepackage{nicefrac}       
\usepackage{microtype}      
\usepackage{xcolor}         

\usepackage{graphicx}
\usepackage{amsmath}
\usepackage{caption}

\usepackage{array}
\usepackage{bm}
\usepackage{multirow}
\usepackage{amssymb}

\def\eg{\emph{e.g.}} 
\def\ie{\emph{i.e.}}

\def\wrt{\emph{w.r.t.}}

\begin{document}

\title{PairAug: What Can Augmented Image-Text Pairs Do for Radiology?}



\author{Yutong Xie$^{1^*}$~~Qi Chen$^{1^*}$~~~Sinuo Wang$^{1}$~~~Minh-Son To$^{4}$~~~Iris Lee$^{4}$~~~Ee Win Khoo$^{4}$\\ 
~~~Kerolos Hendy$^{4}$~~~Daniel Koh$^{4}$~~~Yong Xia$^{2,3}$~~~Qi Wu$^{1^\dagger}$\\
\textit{\normalsize $^{1}$ Australian Institute for Machine Learning (AIML), The University of Adelaide, Australia}\\
\textit{\normalsize $^{2}$ School of Computer Science and Engineering, Northwestern Polytechnical University, China}\\
\textit{\normalsize $^{3}$ Ningbo Institute of Northwestern Polytechnical University, China}\\
\textit{\normalsize $^{4}$ South Australia Medical Imaging, Australia}\\
{\tt\small yutong.xie678@gmail.com, \{qi.chen04, qi.wu01\}@adelaide.edu.au}}






\maketitle
\pagestyle{empty}
\thispagestyle{empty}

\begin{abstract}
Current vision-language pre-training (VLP) methodologies predominantly depend on paired image-text datasets, a resource that is challenging to acquire in radiology due to privacy considerations and labelling complexities. Data augmentation provides a practical solution to overcome the issue of data scarcity, however, most augmentation methods exhibit a limited focus, prioritising either image or text augmentation exclusively. Acknowledging this limitation, our objective is to devise a framework capable of concurrently augmenting medical image and text data.
We design a Pairwise Augmentation (PairAug) approach that contains an Inter-patient Augmentation (InterAug) branch and an Intra-patient Augmentation (IntraAug) branch.
Specifically, the InterAug branch of our approach generates radiology images using synthesised yet plausible reports derived from a Large Language Model (LLM). The generated pairs can be considered a collection of new patient cases since they are artificially created and may not exist in the original dataset.
In contrast, the IntraAug branch uses newly generated reports to manipulate images. This process allows us to create new paired data for each individual with diverse medical conditions.
Our extensive experiments on various downstream tasks covering medical image classification zero-shot and fine-tuning analysis demonstrate that our PairAug, concurrently expanding both image and text data, substantially outperforms image-/text-only expansion baselines and advanced medical VLP baselines. Our code is released at \href{https://github.com/YtongXie/PairAug}{https://github.com/YtongXie/PairAug}.
\end{abstract}  
\renewcommand{\thefootnote}{}
\footnotetext{$^*$Equal contribution. $^\dagger$Corresponding author.}
\section{Introduction}
\label{sec:intro}

Vision-language pre-training (VLP) has garnered considerable attention in recent years~\cite{radford2021learning}, yielding substantial benefits across a wide spectrum of downstream tasks. However, the application of VLP within the medical domain faces a complex challenge, largely attributable to the inherent requirement for extensive data. For instance, the Contrastive Language–Image Pretraining (CLIP) model~\cite{radford2021learning} necessitates training on a dataset comprising 400 million image-text pairs curated from the internet. In stark contrast, the total volume of publicly accessible medical images and reports is significantly lower by several orders of magnitude. This scarcity is primarily due to privacy considerations, data acquisition challenges, and the rarity of certain diseases~\cite{dong2022privacy}.

Data augmentation algorithms serve to address the sample size limit without altering the base model architecture, making them widely applicable across various tasks and algorithms. By expanding medical datasets using such methods, we can not only enhance the size and diversity of training datasets but also impute missing values and maintain patient privacy, thereby reducing dependence on real-world data acquisition.
Many studies~\cite{guibas2017synthetic,shao2023diffuseexpand,pan2021disease,hu2023label,nie2017medical,dalmaz2022resvit} have explored image augmentation techniques designed specifically for medical image expansion, which mainly include traditional spatial transformations and morphological operations, as well as recent image synthesis techniques. These studies demonstrate significant potential in enhancing the accuracy of data-driven diagnostics and prognostics applications.
%
%
Influenced by the recent advancements in Natural Language Processing (NLP), particularly the development of Large Language Models (LLMs)~\cite{touvron2023llama,OpenAI2023GPT4TR,chowdhery2022palm,brown2020language}, many studies have concentrated on the augmentation of medical textual data~\cite{dai2023chataug,tang2023does,yuan2023llm} such as traditional synonym replacement, random deletion and random insertion, as well as more recent methods using LLMs to generate reliable text samples.
%

Despite considerable advances in the field, a significant proportion of medical data augmentation algorithms remain narrowly focused, concentrating exclusively on either image or text augmentation. 
In the context of VLP, image and text data are interconnected and interdependent. Expanding data from a single modality, such as exclusively expanding images or text, fails to fundamentally enhance the information gain. 
This limitation arises from two main factors: Firstly, the augmented modality must retain semantic congruence with the non-augmented one, thereby constraining the scope for diversifying semantic content. For instance, an enlarged X-ray image paired with its original, unchanged report might lead to a description mismatch. Secondly, the information in the non-augmented modality remains unchanged, failing to correspond with modifications in the augmented modality. Consider a scenario where a CT scan image is altered to depict a tumour, but the associated text report remains unaltered, thus not accurately reflecting the changes in the image. Effective augmentation in VLP requires a synchronised enhancement of both image and text, ensuring coherence and maximising the information gain from paired medical image-text data.
We thus argue that the simultaneous expansion of medical image and text data is paramount. By adopting this parallel augmentation strategy, we not only enlarge the dataset quantitatively but also increase data diversity at the semantic level, thereby further enhancing the model's generalisation capability and accuracy.

In this paper, we propose a Pairwise Augmentation (PairAug) approach for medical VLP. The PairAug consists of two distinct branches: the Inter-patient Augmentation (InterAug) and the Intra-patient Augmentation (IntraAug) branches, each with unique functions. These branches aim to expand paired data across inter- and intra-patient domains, maintaining a careful balance to avoid redundancy or overlap within the augmented pairs.
The InterAug branch, powered by large text-to-image models, generates synthetic radiology images from plausible reports produced by the LLM. Given that these image-report pairs are entirely synthesised, they represent an ensemble of novel patient cases, most of which may not exist in the original dataset. These artificially created cases provide us with a new reservoir of data that can augment the current repository of real-world cases.
Conversely, the IntraAug branch operates differently. It modifies existing images according to new reports generated by the model. This unique approach allows us to expand a multitude of new image-report pairs for each individual patient, each pair reflecting a different medical condition. The IntraAug branch thereby allows for creating a diverse dataset that better represents the variety of potential medical scenarios a patient might experience.

We further incorporate two data pruning techniques into our approach to ensure the augmented data's integrity and quality. These methods leverage a pre-trained text-image retrieval model to sift through the data, discarding noisy or irrelevant samples. This rigorous quality control process ensures that our dataset comprises the most relevant and reliable data, making it highly suitable for further analysis and training. 

We conduct medical VLP experiments using real-world pairs from the MIMIC-CXR dataset and PairAug-generated image-report pairs. The learned representations are transferred to classifying diseases under zero-shot and fine-tuning settings on these downstream tasks.
Benefiting from the augmented paired data, our approach achieves mean AUCs of 88.34\% and 70.79\% on the ChestXpert and PadChes zero-shot protocol, respectively surpasses the advanced CheXzero by about 2.10\% and 4.50\% without ensemble. On RSNA fine-tuning protocol, using PairAug-generated pairs beats strong competitors like ImageNet pre-training, and advanced medical image-report pre-training competitor CheXzero. 
Besides, PairAug also outperforms popular image-/text-only augmentation baselines on these downstream tasks. 

\section{Related Works}

\noindent \textbf{Medical VLP}
%
Medical VLP, an extension of VLP in healthcare, aims to interpret complex medical images and associated texts. Many approaches use vision-language contrastive learning~\cite{huang2021gloria, mgca, zhou2022generalized, yan2022clinical,you2023cxr}, leveraging naturally occurring medical image-radiology report pairs, yielding impressive results across various tasks like image classification~\cite{mgca, zhou2022generalized, huang2021gloria} and image-text retrieval~\cite{mgca, huang2021gloria, yan2022clinical}. Recent advancements have seen the incorporation of Masked Auto-Encoder (MAE) model~\cite{he2022masked} from the natural images domain into medical VLP, proving beneficial~\cite{zhou2023advancing, chen2022multi}. However, despite promising outcomes, the performance is often limited by the scarcity and diversity of real-world medical image-text pairs.

\noindent \textbf{Medical Data Expansion}
The field of medical data expansion is an active area of research, addressing the critical issue of data scarcity in healthcare~\cite{kebaili2023deep}. 
Traditional image expansion methods have primarily adopted spatial transformations such as rotation, scaling, and flipping, as well as morphological operations like cropping and padding~\cite{zhao2019data,xie2021cotr,isensee2021nnu}. Although these techniques are straightforward to implement, they may fall short of capturing the intricate variations inherent in medical images.
The emergence of advanced augmentation techniques, particularly Generative Adversarial Networks (GANs), has ushered in a new era of synthetic but realistic-looking image generation. GANs and their variants have demonstrated considerable success across various medical imaging contexts~\cite{pan2021disease,yi2019generative,nie2017medical}, offering a richer, more diverse dataset for model training.
%
A handful of recent works have explored the potential of Language-to-Image models in medical image expansion~\cite{rajapaksa2023using,chambon2022roentgen,yang2020xraygan,salehinejad2018generalization}. They employ prompts based on medical textbooks/reports as inputs for generating images and improving the diagnostic accuracy of models trained on limited real datasets.

Traditional text expansion methods work at different granularity levels~\cite{dai2023chataug}: characters, words, sentences, and documents. Recent advances in LLMs, \eg, PaLM~\cite{chowdhery2022palm}, LLaMA~\cite{touvron2023llama}, and ChatGPT, have facilitated the development of more sophisticated text expansion methods.
%
Despite substantial progress in medical data expansion, most existing strategies are confined to either image or text synthesis. This singular focus does not adequately address the persistent issue of limited real-world medical image-text pair data.


\section{Proposed Method}

\begin{figure*}[t]
    \centering
    {
    \includegraphics[width=0.89\linewidth]{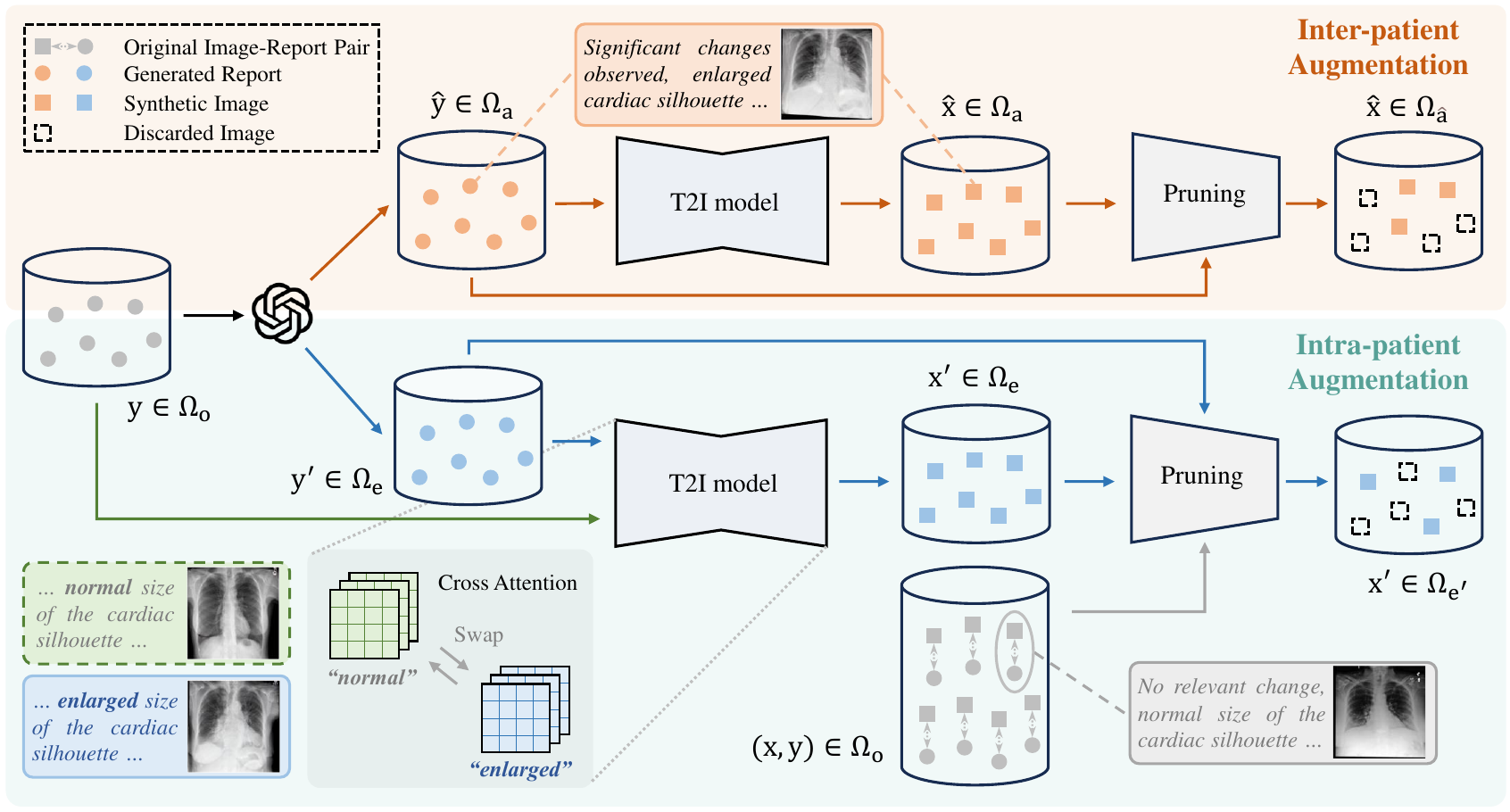}
    }
    \vspace{-0.2cm}
    \caption{Overall of Pairwise Augmentation (PairAug) pipeline, consisting of two branches: Inter-patient Augmentation (InterAug) and Intra-patient Augmentation (IntraAug). 
    In InterAug, we first generate new reports $\hat{y}\in\Omega_a$ by a large language model $\mathcal{P}$ from original reports $y\in\Omega$. Then, we synthesise images $\hat{x}\in\Omega_a$ from the generated reports, followed by a data pruning method \wrt~the semantic alignment between generated image-report pairs.
    As for IntraAug, we seek to generate images for the same individual but with different medical conditions. To this end, we reuse the same generation model $\mathcal{G}$ to synthesise images but swap the cross-attention map $M$ from the original report $y$ with that (\ie, $M'$) from the modified report $y'$ during the generation process. After that, we consider a data pruning method based on both synthetic pairs $(x',y')\in\Omega_e$ and original pairs $(x,y)\in\Omega$.
    Last, we merge $\Omega_{\hat{a}}$ and $\Omega_{e'}$ as the final synthetic paired data set $\Omega_{\tilde{s}}$.
    }
    \label{fig:overall}
    \vspace{-0.3cm}
\end{figure*}

%


\noindent \textbf{Problem Statement}
To tackle the issue of limited data, we explore a novel dataset augmentation task. Considering open-vocabulary learning, we specifically aim to augment image-text pairs in an original/existing dataset $\Omega_o$, where $(x_i, y_i)\in\Omega_o$. Here, $x_i$ represents an image paired with corresponding text $y_i$ ($e.g.,$ radiology reports), and $n_o$ indicates the number of samples. The objective of dataset expansion is to generate a collection of new synthetic samples $\Omega_s$, where $(\tilde{x}_i, \tilde{y}_i)\in\Omega_s$, to amplify the original dataset, enabling a learnable model trained on the expanded dataset $\Omega_o\cup\Omega_s$ significantly outperform a model trained solely on $\Omega_o$.
Crucially, the synthetic pairs set $\Omega_s$ should bring sufficient new and correct information to boost the model's training.

\noindent \textbf{How to Augment for Effective Expansion?}
To generate new image-text pairs, we leverage the capabilities of the large language model $\mathcal{P}$ and the image synthesis model $\mathcal{G}$, known for their impressive text and image generation capabilities, respectively. 
However, the effectiveness of different sample types remains unclear. Our main insight is that the newly synthesised pairs $(\tilde{x}, \tilde{y}) \in \Omega_s$ should introduce new information compared to the original pairs $(x, y) \in \Omega_o$ while maintaining a high quality for each created pair. To achieve these, we consider two key criteria: (1) non-overlapped pairwise augmentation and (2) prioritising high-quality pairs.

\noindent \textbf{Overall Pipeline}
As shown in Figure~\ref{fig:overall}, considering the aforementioned points, we propose a framework called Pairwise Augmentation (PairAug) for expanding datasets.
This framework, guided by the specified criteria, broadens the dataset through two distinct branches: Inter-patient Augmentation (InterAug) and Intra-patient Augmentation (IntraAug), thereby preventing redundancy or overlap in the augmented pairs.
Moreover, we incorporate specific data pruning methods for each branch to uphold the quality of the augmented pairs.
The pipeline of PairAug can be formulated as
%
%
\begin{equation}
    \small
    \begin{aligned}
        \Omega_{\tilde{s}} \leftarrow \texttt{Pr}(\Omega_s),~\text{s.t.}~\Omega_s=\{(\tilde{x}_i, \tilde{y}_i)|\tilde{x}_i=\mathcal{G}(\tilde{y}_i), \tilde{y}_i=\mathcal{P}(y_i)\}_{i=1}^{n_s},
    \end{aligned}
    \label{eq:overall}
\end{equation}
where $\Omega_{\tilde{s}}$ is the subset of $\Omega_s$ and $n_s$ denotes the number of synthetic pairs in $\Omega_s$. $\texttt{Pr}()$ is a pruning operation. For simplicity, we omit the input prompt for large language model $\mathcal{P}$.
In practice, the final synthetic data set $\Omega_{\tilde{s}}$ consists of two subsets $\Omega_{\hat{a}}$ and $\Omega_{e'}$ derived from our InterAug and IntraAug branches, respectively.
In the following, we will depict how to obtain $\Omega_{\hat{a}}$ and $\Omega_{e'}$, as well as the specific formulation of Eq.~(\ref{eq:overall}) in the proposed two branches.
%
%


\begin{table}[t]
    \centering
    \captionof{table}{Reports before and after modifying based on our prompt. First row: abnormal $\rightarrow$ normal; second row: normal $\rightarrow$ abnormal.}
    \vspace{-0.3cm}
    \resizebox{1.0\linewidth}{!}{
    \begin{tabular}{p{0.53\linewidth}|p{0.47\linewidth}}
    \toprule
    Report (before) & Report (after) \\
    \midrule
        \textit{Mild pulmonary edema with superimposed left upper lung consolidation...} & \textit{No pulmonary edema or lung consolidation is observed...} \\
        \midrule
        \textit{...lungs are hyperinflated though clear, cardio mediastinal silhouette is stable, ...} & \textit{...lungs are collapsed and unclear, cardio mediastinal silhouette is unstable, ...} \\
    \bottomrule
    \end{tabular}
    }
    \label{tab:report_example} 
\end{table}

\subsection{InterAug: Inter-patient Augmentation}

\noindent \textbf{New Report Generated by Large Language Model (LLM)}
Firstly, we focus on the text domain, using ChatGPT with Azure OpenAI service to process X-ray reports. The LLM produces the resulting report based on manual instruction~(prompt).
To simplify the process, we aim to find a single satisfactory prompt when generating new reports, \ie, ``\textit{Following is an original chest X-Ray report. Generate one possible augmentation that is limited to 50 words while conveying partial opposite meanings than the original report}''.
%

As shown in Table~\ref{tab:report_example}, we take as an example that provided the input report ``\textit{Mild pulmonary edema with superimposed left upper lung consolidation}'', we use ChatGPT to generate an appropriately modified output report ``\textit{No pulmonary edema or lung consolidation is observed}'' based on our written prompt.
In this way, we can obtain a large number of new and diverse radiology reports.
Due to the page limit, we provide more examples in the supplementary.

\begin{figure}[t]
    \centering 
    {
    \includegraphics[width=1.0\linewidth]{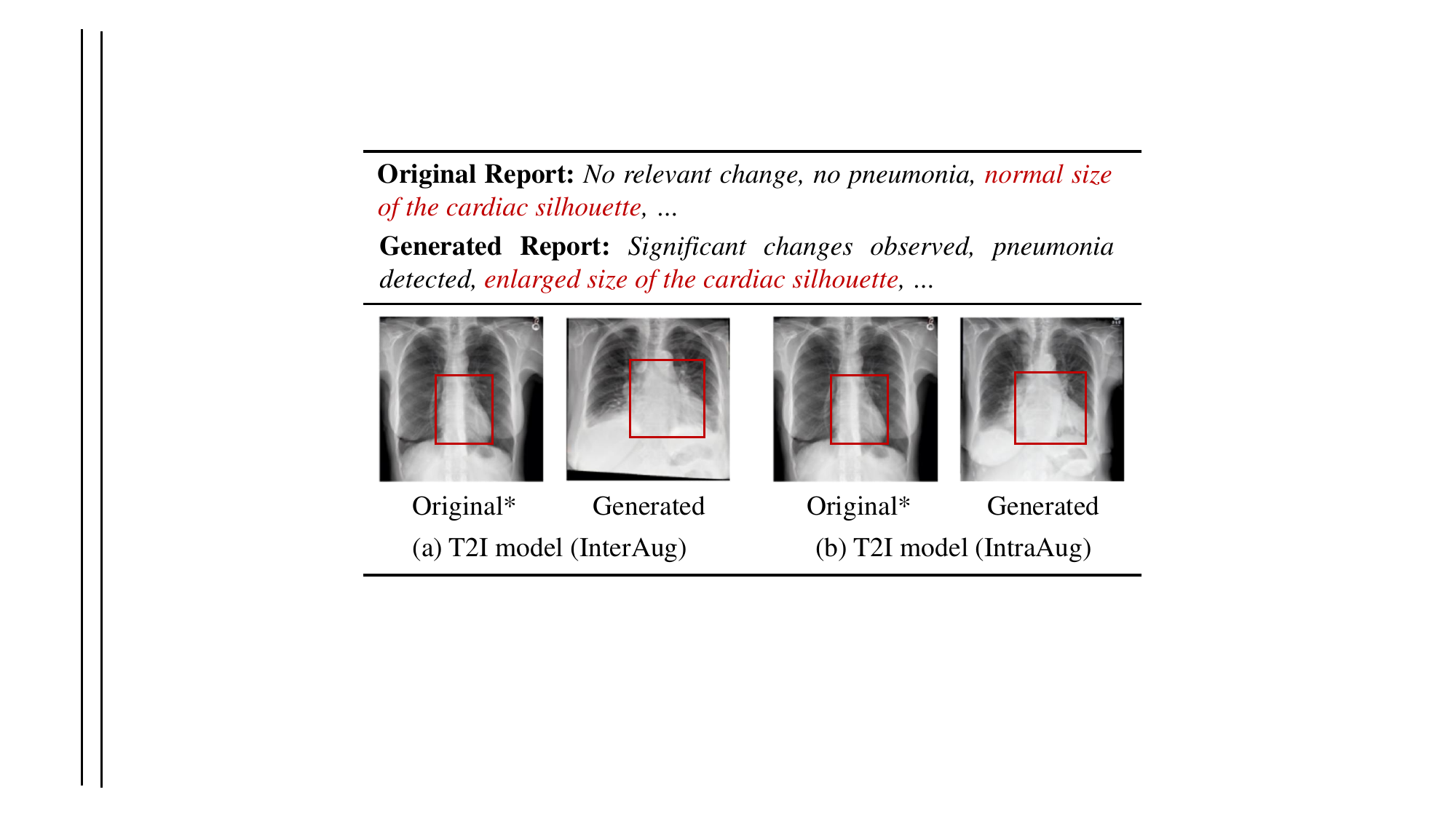} 
    }
    \vspace{-0.7cm}
    \caption{Images synthesised from original and generated reports by (a) the T2I model in InterAug and (b) the T2I model in IntraAug with attention map swapping, respectively. * denotes images generated from the original reports rather than the original images.}
    \label{fig:image_example_p2p} 
    \vspace{-0.3cm}
\end{figure}

\noindent \textbf{Inter-patient Image Generation}
In this part, we seek to generate images based on text without any other constraints. In this way, these pairs can be regarded as a set of new patients as they are synthetically generated and may not exist in original datasets (see the generated image in Figure~\ref{fig:image_example_p2p}(a)).
Specifically, we base our model on Stable Diffusion, a large-scale text-to-image (T2I) latent diffusion model~\cite{rombach2022high} and use the post-pretraining version on radiology datasets, namely RoentGen~\cite{chambon2022roentgen}. Formally, the process of generating new image-text pairs can be defined as
\begin{equation}
    \begin{aligned}
        \Omega_a=\{(\hat{x}_i, \hat{y}_i)|\hat{x}_i=\mathcal{G}_{z\sim p_z}(\hat{y}_i, z), \hat{y}_i = \mathcal{P}_{y_i\in\Omega_o}(y_i)\}_{i=1}^{n_a},
    \end{aligned}
\end{equation}
where $\mathcal{P}$ refers to the LLM and $\mathcal{G}$ is the generation model. $p_z$ is the prior distribution for input noise variable $z$. $n_a$ means the number of new image-text pairs.

\noindent \textbf{Data Pruning by Semantically-aligned Informativeness} 
To ensure the quality of the generated pairs, we introduce a data pruning method $\texttt{Pr}_{a}()$ based on a semantically-aligned score $\mathcal{S}_{a}$, which resorts to the semantic alignment abilities of CLIP~\cite{radford2021learning}.
Similar to~\cite{ramesh2021zero}, we rerank the samples drawn from the generation model using CLIP. Notably, as we focus on the radiology image generation in chest X-rays, we employ MedCLIP~\cite{wang2022medclip} pre-trained on the chest X-ray dataset rather than the original CLIP model.
Specifically, for the paired data from our InterAug, MedCLIP assigns a score based on how well the image matches the report. 
Then, we filter and only retain those image-report pairs that have attained scores exceeding the threshold $\tau$.
In this way, we not only ensure the semantically aligned informativeness of the extended data set but also improve the robustness of our data generation method against failures of the generation model.
Mathematically, the retained data set $\Omega_{\hat{a}}$ after pruning process can be defined as
%
\begin{equation}
    \begin{aligned}
     \Omega_{\hat{a}} \leftarrow \texttt{Pr}_{a}(\Omega_a, \tau) = \{(\hat{x},\hat{y})| (\hat{x}, \hat{y})\in \Omega_a, \mathcal{S}_{a}(\hat{x},\hat{y}) > \tau\},
    \end{aligned}
\end{equation}
 where $\mathcal{S}_{a}(\hat{x},\hat{y})$ refers to the cosine similarity between the image feature and text feature extracted by MedCLIP's image encoder and text encoder, respectively.

\subsection{IntraAug: Intra-patient Augmentation}
Besides expanding the data at the patient level through InterAug, we seek to obtain samples that capture various medical conditions for each individual. This further enhances the diversity of dataset without introducing overlap or redundancy compared to data generated by InterAug.
%
%
However, due to the lack of guarantees regarding image consistency, capturing changes in each patient's condition, even minor changes, is challenging.
In Figure~\ref{fig:image_example_p2p}(a), the radiology images are from different patients, even if both of them are semantically aligned with their reports (see the box and description in red).
This contradicts our goal, as we hope to generate radiology images displaying different medical conditions for the same patient (Figure~\ref{fig:image_example_p2p}(b)), rather than generating conditions on another patient.
%
To this end, we design an IntraAug to yield new intra-patient images using newly generated reports while considering the original reports as conditions. 
%
%
To make the whole model memory-friendly, instead of re-developing an image editing model (in which the editing quality may not be guaranteed with limited training data), we reuse the same generation model $\mathcal{G}$ (from the above section) to synthesise images but swap the cross-attention map from the original report. See the below section for more details.

%

%
%

\noindent \textbf{Image Generation by Controlling Cross-Attention Maps} 
Inspired by the observation in~\cite{hertz2022prompt}, 
we seek to control the generation process by modifying the intermediate cross-attention maps $M$ of generation model $\mathcal{G}$.
%
Formally, we define the generation process as
\begin{equation}
    \begin{aligned}
        \Omega_e=\{(x'_i, y'_i)|x'_i=\mathcal{E}(y_i, y'_i), y'_i = \mathcal{P}_{y_i\in\Omega_o}(y_i)\}_{i=1}^{n_e},
    \end{aligned}
\end{equation}
where $\mathcal{P}$ is the large language model and $n_e$ means the total number of generated data. Here, $\mathcal{E}$ refers to the generation process, which contains the diffusion model $\mathcal{G}$ and an attention map swapping operation $\texttt{Swap}^{(t)}(M_t, M'_t)$.
This operation swaps the original cross-attention maps $M_t$ (from the original report $y$) with the modified $M'_t$ (from the modified report $y'$) at step $t$ of the diffusion process, \eg, from \textit{``normal''} to \textit{``enlarged''} in Figure~\ref{fig:overall}.

Mathematically, given $y'$, the output noisy image $z'_{t-1}$ at the step $t$ of diffusion processing can be calculated by
\begin{equation}
    \begin{aligned}
        z'_{t-1} \leftarrow \mathcal{E}^{(t)}(y, y') = \mathcal{G}^{(t)}(y',z'_t)\{\texttt{Swap}^{(t)}(M_t, M'_t)\}.
    \end{aligned}
\end{equation}
Here, let $\mathcal{G}^{(t)}(y',z'_t)\{\texttt{Swap}^{(t)}(M_t, M'_t)\}$ represent the diffusion step where we swap the attention map $M_t$ with the modified map $M'_t$, where $M_t\leftarrow\mathcal{G}^{(t)}(y, z_t)$ and $M'_t\leftarrow\mathcal{G}^{(t)}(y', z'_t)$. The noisy images $z_t$ and $z'_t$ are generated at the previous step from $y$ and $y'$, respectively. For $\mathcal{G}$, a random noise $z'_T\sim p_z$ is fed at the first step $T$ and finally yields the image $x'=z'_0$ at the last step $0$.
%
Moreover, following~\cite{hertz2022prompt,liu2023video}, we use a softer attention map swapping method for well-controlling the degree of modification, \ie,
\begin{equation}
    \begin{aligned}
        \texttt{Swap}^{(t)}(M_t, M'_t):= \left\{
        \begin{array}{ccl}
             M_t \leftarrow M'_t & & t < \eta \\
             M_t & & \text{otherwise},  \end{array}
        \right.
    \end{aligned}
\end{equation}
where $\eta=0.5$ is a timestamp hyper-parameter, specifying which step the swapping operation is used. 

\noindent \textbf{Data Pruning by Hybrid Consistency}
%
In this part, we assess the data from IntraAug by designing a hybrid consistency score, which consists of three consistency criteria: (i) semantic alignment $\mathcal{S}_{1}$ between input reports and corresponding images; (ii) similarity $\mathcal{S}_{2}$ between original images and generated images; (iii) the consistency of the change $\mathcal{S}_{3}$ between two images with the change between the corresponding two reports.
For (i) and (ii), we directly use the MedCLIP to capture both features of reports and images, and then calculate the similarity for image-report pairs and image-image pairs, respectively.
As for (iii), inspired by~\cite{gal2022stylegan}, we use the directional similarity in CLIP space. This calculates the cosine similarity between $\Delta x$ and $\Delta y$, where $\Delta x$ represents the differences between image features and $\Delta y$ represents the differences between report features.

Due to the different magnitudes of $\mathcal{S}_1$, $\mathcal{S}_2$, and $\mathcal{S}_3$, we seek to computer them individually and subsequently take their mean values as filtering thresholds.
Formally, the data sets $\Omega_1$, $\Omega_2$ and $\Omega_3$ can be filtered by
\begin{equation}
    \small
    \left\{
    \begin{array}{lr}
     \Omega_1 = \{(x',y') | \mathcal{S}_1(x',y')>(\mu_1-\epsilon)\} & \\
     \Omega_2 = \{(x',y') | \mathcal{S}_2(x',x)>(\mu_2-\epsilon)\} & \\
     \Omega_3 = \{(x',y') | \mathcal{S}_3(\Delta x,\Delta y)>(\mu_3-\epsilon)\}.
     \end{array}
    \right.
\end{equation}
Here, $\Delta x$ is the subtraction of features between $x$ and $x'$ while $\Delta y$ is that between $y$ and $y'$, where $(x', y')\in \Omega_e$ and $(x, y)\in \Omega_o$. The $\Omega_o$ is the original data set. $\mu_1$, $\mu_2$ and $\mu_3$ denote the mean value of score $\mathcal{S}_1$, $\mathcal{S}_2$ and $\mathcal{S}_3$, respectively.
Besides, we introduce a hyper-parameter $\epsilon$, making the threshold more flexible and less stringent.
Finally, we obtain the augmented dataset by
%
\begin{equation}
    \begin{aligned}
     \Omega_{e'} \leftarrow \texttt{Pr}_{e}(\Omega_e, \Omega_o, \epsilon) = \Omega_1 \cap \Omega_2 \cap \Omega_3,
    \end{aligned}
\end{equation}
where $\texttt{Pr}_e()$ is the pruning method with above three criteria.


\subsection{Medical VLP with Generated Pairs}


We amalgamate real-world and generated medical image-text pairs for our medical VLP. Our model builds upon the CheXzero framework~\cite{tiu2022expert}, an advanced approach that proficiently exploits semantic correspondences between medical images and radiology reports for comprehensive medical data representation learning. 
We use the Vision Transformer~\cite{dosovitskiyimage}, ViT-B/32, as the image encoder and employ a Transformer~\cite{vaswani2017attention} with 12 layers and a width of 512 with eight attention heads for the text encoder. We initialized the self-supervised model using the pre-trained weights from OpenAI’s CLIP model~\cite{radford2021learning}.
After that, we apply the pre-trained weight parameters to various downstream classification tasks under zero-shot and fine-tuning settings. For the zero-shot setting, we followed the CheXzero and used a positive-negative softmax evaluation procedure on each disease for multi-label classification task. In particular, we compute logits with positive prompts (such as pneumonia) and negative prompts (that is, no pneumonia). Then, we compute the softmax between the positive and negative logits. Lastly, we keep the softmax probabilities of the positive logits as the probability of the disease in the chest X-ray.
We employ the widely-used linear probing approach for the fine-tuning evaluation, in which the pre-trained image encoder is frozen, and only a randomly initialized linear classification head is trained.  

\section{Experiments}

\subsection{Implementation Details}

\noindent \textbf{Generation Setup}
Following~\cite{chambon2022roentgen}, we establish a guidance scale of 4 and generate images at 512 resolution with 75 denoising steps using a PNDM noise scheduler~\cite{liu2022pseudo}. 
For data pruning, we empirically set the $\tau$ and $\epsilon$ thresholds to 0.3 and 0.003, respectively, to guarantee the quality of generated image-text pairs.
Finally, our PairAug generates 187,922 image-text pairs; 44,279 are from InterAug, while IntraAug produces the remaining 143,643 pairs.

\noindent \textbf{Pre-training Setup}
%
We combine real-world 377k imge-report pairs from the MIMIC-CXR dataset~\cite{johnson2019mimic} and PairAug-generated image-report pairs for the pre-training. 
During the pre-training phase, we combine the real-world and generated pairs. We set the input image size at $224\times224$ and normalise each image using the training dataset's sample mean and standard deviation. We tokenise reports using byte pair encoding with a 49,408-word vocabulary. 
%
We use the stochastic gradient descent optimiser combined with a learning rate of 0.0001, a momentum of 0.9, and a batch size of 64. The maximum training epoch is ten.

\noindent \textbf{Downstream Setup}
We test the performance of learned VLP representations on three radiology-based downstream datasets:
(1) CheXpert dataset~\cite{irvin2019chexpert} contains 191,229 frontal chest radiographs. We use its official test set for zero-shot evaluation aiming to classify each image into 5 five individual binary labels: atelectasis, cardiomegaly, consolidation, edema, and pleural effusion; 
(2) PadChest dataset~\cite{bustos2020padchest} has 193 disease image labels, including 174 radiographic findings and 19 differential diagnoses. We adopt 39,053 chest X-rays annotated by board-certified radiologists for zero-shot evaluation;
(3) pneumothorax classification on the RSNA Pneumonia dataset~\cite{shih2019augmenting} with over 29k frontal view chest radiographs, which aims to classify each radiograph into negative or positive for pneumothorax. We split the dataset into training, validation, and test sets with an 80\%/10\%/10\% ratio.
We report the area under the ROC curve (AUC), accuracy (ACC) and macro-averaged F1 score (F1) as the evaluation metrics. We put more details in the supplementary.

\subsection{Comparison with State-of-the-arts}

We compare our PairAug with different pre-training and data augmentation methods on three downstream datasets, including zero-shot evaluation on ChestXpert and PadChest datasets in Table~\ref{tab:zeroshot} and linear probing evaluation on RSNA datasets under varying ratios of available labelled data (1\%, 10\%, and 100\%) in Table~\ref{tab:linear}.
The compared methods include ImageNet pre-training~\cite{dosovitskiyimage}, popular medical image-report pre-training approaches MGCA~\cite{huang2021gloria}, CXR-CLIP~\cite{you2023cxr} and CheXzero (our base model)~\cite{tiu2022expert}. 
Besides, we test the impact of our base model with only image augmentation (Base+AugImg), with only text augmentation (Base+AugText), with image+text augmentation (Base+AugText+AugImg), and with our paired image-text data augmentation (Base+PairAug).
For AugImg, we incorporate traditional image augmentation in VLP, \ie, random cropping, Gaussian Blur, and Grayscale, to expand the image data.
For AugText, we employ ChatGPT to rewrite the original reports while keeping the same semantics, to expand the text data.
For AugText+AugImg, we add traditional image augmentation to the AugText setting.

\begin{table*}[t]
\caption{Classification results (AUC, ACC and F1) of different pre-training methods on two downstream test sets under zero-shot evaluation. 
For ChestXpert, the metrics all are the average of five diseases. For PadChest, the metrics all are the average of 193 diseases. Numbers within parentheses indicate 95\% confidence interval (CI). `S' denotes a single model. `E' is the ensemble over top-ten model checkpoints.}
\vspace{-0.6cm}
\label{tab:zeroshot}
\begin{center}
\resizebox{1.0\linewidth}{!}
{
\begin{tabular}{c|cccccc}
\hline
\multirow{2}{*}{Methods} & \multicolumn{3}{c|}{ChestXpert}                                       & \multicolumn{3}{c}{PadChest}                                        \\ \cline{2-7} 
                         & \multicolumn{1}{c|}{AUC(\%)}     & \multicolumn{1}{c|}{Acc(\%)}     & \multicolumn{1}{c|}{F1(\%)}      & \multicolumn{1}{c|}{AUC(\%)}     & \multicolumn{1}{c|}{Acc(\%)}     & F1(\%)     \\ \hline
MGCA (NeurIPS'22)        & \multicolumn{1}{c|}{84.29\footnotesize{(79.88, 88.33)}} & \multicolumn{1}{c|}{82.11\footnotesize{(76.04, 87.00)}} & \multicolumn{1}{c|}{61.12\footnotesize{(53.00, 69.00)}} & \multicolumn{1}{c|}{66.12\footnotesize{(59.56, 72.05)}} & \multicolumn{1}{c|}{81.38\footnotesize{(63.32, 91.68)}} & 4.89\footnotesize{(4.13, 6.60)} \\ 
CXR-CLIP (MICCAI'23)        & \multicolumn{1}{c|}{86.20\footnotesize{(82.02, 90.08)}} & \multicolumn{1}{c|}{83.24\footnotesize{(76.28, 88.20)}} & \multicolumn{1}{c|}{61.63\footnotesize{(53.44, 69.73)}} & \multicolumn{1}{c|}{68.50\footnotesize{(61.87, 74.52)}} & \multicolumn{1}{c|}{86.04\footnotesize{(71.85, 94.13)}} & \textbf{8.62}\footnotesize{(6.63, 11.57)} \\ 
CheXzero-E (Nat.BE'22)    & \multicolumn{1}{c|}{88.92\footnotesize{(84.97, 92.29)}} & \multicolumn{1}{c|}{85.75\footnotesize{(81.84, 89.44)}} & \multicolumn{1}{c|}{66.51\footnotesize{(58.62, 73.56)}} & \multicolumn{1}{c|}{71.24\footnotesize{(65.52, 76.05)}} & \multicolumn{1}{c|}{84.54\footnotesize{(70.50, 92.45)}} & 7.36 \footnotesize{(5.31, 9.34)} \\ 
CheXzero-S (Base)        & \multicolumn{1}{c|}{86.24\footnotesize{(81.77, 90.16)}} & \multicolumn{1}{c|}{83.13\footnotesize{(74.76, 88.24)}} & \multicolumn{1}{c|}{59.98\footnotesize{(52.47, 67.18)}} & \multicolumn{1}{c|}{66.29\footnotesize{(59.57, 72.23)}} & \multicolumn{1}{c|}{79.44\footnotesize{(64.22, 90.69)}} & 6.15\footnotesize{(5.00, 8.02)} \\ \bottomrule
Base + AugImg            & \multicolumn{1}{c|}{87.43\footnotesize{(83.37, 90.96)}} & \multicolumn{1}{c|}{84.07\footnotesize{(78.48, 88.44)}} & \multicolumn{1}{c|}{62.16\footnotesize{(54.55, 69.29)}} & \multicolumn{1}{c|}{68.12\footnotesize{(61.74, 73.96)}} & \multicolumn{1}{c|}{83.73\footnotesize{(68.44, 92.56)}} & 6.15\footnotesize{(4.87, 8.25)} \\ 
Base + AugText           & \multicolumn{1}{c|}{87.03\footnotesize{(82.72, 90.55)}} & \multicolumn{1}{c|}{82.43\footnotesize{(77.24, 87.16)}} & \multicolumn{1}{c|}{62.07\footnotesize{(54.52, 69.11)}} & \multicolumn{1}{c|}{67.35\footnotesize{(60.82, 73.14)}} & \multicolumn{1}{c|}{81.93\footnotesize{(66.64, 90.65)}} & 6.41\footnotesize{(5.13, 8.49)} \\ 
Base + AugText + AugImg  & \multicolumn{1}{c|}{87.12\footnotesize{(82.94, 90.86)}} & \multicolumn{1}{c|}{83.69\footnotesize{(77.96, 88.40)}} & \multicolumn{1}{c|}{62.46\footnotesize{(54.64, 69.99)}} & \multicolumn{1}{c|}{69.23\footnotesize{(63.01, 74.48)}} & \multicolumn{1}{c|}{83.39\footnotesize{(68.37, 92.46)}} & 5.91\footnotesize{(4.63, 8.10)} \\ \hline
Base + PairAug (Ours)    & \multicolumn{1}{c|}{88.34\footnotesize{(84.31, 91.84)}} & \multicolumn{1}{c|}{84.97\footnotesize{(80.00, 88.76)}} & \multicolumn{1}{c|}{65.73\footnotesize{(57.65, 73.11)}} & \multicolumn{1}{c|}{70.79\footnotesize{(64.90, 75.97)}} & \multicolumn{1}{c|}{84.90\footnotesize{(71.63, 93.18)}} & 7.51\footnotesize{(5.99, 9.95)} \\ 
Base + PairAug-E (Ours)  & \multicolumn{1}{c|}{\textbf{89.97}\footnotesize{(86.00, 93.27)}}        & \multicolumn{1}{c|}{\textbf{86.21}\footnotesize{(81.60, 89.92)}}        & \multicolumn{1}{c|}{\textbf{67.78}\footnotesize{(59.72, 75.07)}}        & \multicolumn{1}{c|}{\textbf{72.51}\footnotesize{(66.36, 77.77)}}        & \multicolumn{1}{c|}{\textbf{86.51}\footnotesize{(72.59, 93.77)}}        &  7.67 \footnotesize{(6.28, 10.08)} \\   
\bottomrule
\end{tabular}
}
\end{center}
\vspace{-0.6cm}
\end{table*}

\begin{table}[t]
\caption{Classification results of different pre-training methods on RSNA test sets under different ratios of available labelled data.}
\vspace{-0.6cm}
\label{tab:linear}
\begin{center}
\resizebox{1.0\linewidth}{!}
{
\begin{tabular}{c|ccc|ccc}
\hline
\multirow{2}{*}{Methods} & \multicolumn{3}{c|}{1\%}                                        & \multicolumn{3}{c}{100\%}                                      \\ \cline{2-7} 
                         & \multicolumn{1}{c|}{AUC}   & \multicolumn{1}{c|}{Acc}   & F1    & \multicolumn{1}{c|}{AUC}   & \multicolumn{1}{c|}{Acc}   & F1    \\ \hline
Rand                     & \multicolumn{1}{c|}{61.48} & \multicolumn{1}{c|}{77.62} & 43.70  & \multicolumn{1}{c|}{77.67} & \multicolumn{1}{c|}{79.01} & 59.89 \\ 
ImageNet                 & \multicolumn{1}{c|}{74.98} & \multicolumn{1}{c|}{77.51} & 48.53 & \multicolumn{1}{c|}{83.63} & \multicolumn{1}{c|}{81.60}  & 69.19 \\ 
CheXzero-S (Base)        & \multicolumn{1}{c|}{83.45} & \multicolumn{1}{c|}{80.5}  & 70.05 & \multicolumn{1}{c|}{86.99} & \multicolumn{1}{c|}{83.56} & 75.08 \\ \bottomrule
Base + AugImg            & \multicolumn{1}{c|}{84.80}  & \multicolumn{1}{c|}{82.02} & 72.28 & \multicolumn{1}{c|}{87.84} & \multicolumn{1}{c|}{84.67} & 76.41 \\ 
Base + AugText           & \multicolumn{1}{c|}{84.42} & \multicolumn{1}{c|}{81.76} & 71.83 & \multicolumn{1}{c|}{87.40}  & \multicolumn{1}{c|}{83.82} & 75.31 \\ 
Base + AugText + AugImg  & \multicolumn{1}{c|}{84.79} & \multicolumn{1}{c|}{81.93} & 71.79 & \multicolumn{1}{c|}{87.78} & \multicolumn{1}{c|}{84.68} & 76.22 \\ 
Base + PairAug (Ours)    & \multicolumn{1}{c|}{\textbf{85.89}} & \multicolumn{1}{c|}{\textbf{83.13}} & \textbf{73.51} & \multicolumn{1}{c|}{\textbf{88.63}} & \multicolumn{1}{c|}{\textbf{85.23}} & \textbf{77.10}  \\ \hline
\end{tabular}
}
\end{center}
\vspace{-0.6cm}
\end{table}

\noindent \textbf{Comparison with Different Augmentation Methods}
When comparing with image-only and text-only data augmentation methods (\ie, AugImg, AugText, and their combination), Table~\ref{tab:zeroshot} shows that our PairAug exhibits superior performance across all these scenarios.

Although AugImg and AugText contribute to performance improvement compared to the base model (CheXzero-S), their data augmentation focuses on only a single modality (either image or text), under the premise of semantic invariance, limiting their potential to enrich the information gain fundamentally. 
Besides, limited to the data diversity at the semantic level, it is interesting that performance is not significantly enhanced even when these two augmentation techniques are combined.
In contrast, our PairAug approach generates paired medical image-text data, augmenting both image and text modalities concurrently, and importantly, without the restrictions of semantic invariance. Thus, PairAug produces a more comprehensive and contextually rich training dataset for the model, which, as demonstrated by the experimental results, improves model performance.

For ChestXpert and PadChest datasets, the performance gain achieved by PairAug over AugImg, AugText, and AugImg+AugText is significant for all metrics with a 95\% confidence interval (CI). 
E.g., for ChestXpert, PairAug achieves an average AUC of 88.34\%, an average AUC of 84.97\%, and an F1 score of 65.73\%, which are higher than the scores achieved by AugImg (87.43, 84.07 and 62.16), AugText (87.03, 82.43 and 62.07), and AugImg+AugText (87.12, 83.69 and 62.46). 
The total minimum performance gain over three augmentation methods is 5.08\% on the ChestXpert dataset and 3.82\% on the PadChest dataset.
On the RSNA dataset, PairAug consistently outperforms three augmentation methods across all proportions of available labelled data. Even with only 1\% of labelled data, PairAug achieves an AUC of 85.89\%, an accuracy of 83.13\% and an F1 score of 73.51\%. 
This consistent outperformance of PairAug over all these augmentation methods across different scenarios illustrates the importance and effectiveness of generating paired image-text data for medical VLP tasks.


\noindent \textbf{Comparison with Different Pre-training Methods}
In Table~\ref{tab:zeroshot}, our PairAug, using CheXzero as a base model, sets new state-of-the-art zero-shot learning benchmarks for two key downstream datasets. Our method also beats the CheXzero model with or without using ensembles.
For the fine-tuning comparisons, Table~\ref{tab:linear} shows that compared to models trained on ImageNet only, those further pre-trained on medical datasets—specifically CheXzero—consistently perform better, regardless of the labelling ratio. This demonstrates the importance of tailored pre-training for medical imaging, which has unique characteristics and requirements.
Moreover, our PairAug consistently outperforms CheXzero on the RSNA dataset whenever using different proportions of labelled medical data for fine-tuning. 
These superior results further suggest the effectiveness of our data augmentation strategy.
By augmenting image and text data concurrently, PairAug enhances the generalisation performance of these downstream tasks. 
This is particularly beneficial when only a limited amount of labelled data is available.


\subsection{Ablation Study}\label{sec:ablation}

To test the performance of each component in PairAug, we conduct an ablation study for InterAug, IntraAug and two data pruning strategies in Table~\ref{tab:ablation}. We gradually add each component to the base model to observe downstream zero-shot performance trends on ChestXpert and PadChest datasets.
First, incorporating the InterAug/IntraAug module leads to a limited, even decreased, performance gain on both datasets. This can be attributed to the generated data that possibly introduce noise or inconsistencies, causing the model to underperform.
The subsequent addition of both data pruning mechanisms results in a noticeable improvement in average performance (+3.05\%) compared with the base model, demonstrating the value of this component in filtering out low-quality synthesised data and reducing noise in the training set.
Finally, when we incorporate all components into the framework, the performance on both datasets reaches its peak (+5.06\%). 
It suggests that jointly two branches can provide more diverse data for pre-training, further enhancing the model's generalisation ability.

%
\begin{table}[t]
\caption{Ablation study. $\Delta$ is average performance gain compared to Base. `ChestX.': CheXpert dataset. `PadC.': PadChest dataset.}
\vspace{-0.6cm}
\label{tab:ablation}
\begin{center}
\resizebox{1.0\linewidth}{!}
{
\begin{tabular}{ccccc|c|c|c}
\hline
\multicolumn{5}{c|}{Ablations}                                                                                                              & ChestX. & PadC. & \multirow{2}{*}{$\Delta$} \\ \cline{1-7}
\multicolumn{1}{c|}{Base} & \multicolumn{1}{c|}{InterAug} & \multicolumn{1}{c|}{$\texttt{Pr}_{a}$} & \multicolumn{1}{c|}{IntraAug} & $\texttt{Pr}_{e}$ & AUC        & AUC      &                                        \\ \hline
\multicolumn{1}{c|}{$\checkmark$}    & \multicolumn{1}{c|}{}         & \multicolumn{1}{c|}{}            & \multicolumn{1}{c|}{}          &             & 86.24    & 66.29  & -                                      \\ \hline
\multicolumn{1}{c|}{$\checkmark$}    & \multicolumn{1}{c|}{$\checkmark$}        & \multicolumn{1}{c|}{}            & \multicolumn{1}{c|}{}          &             & 84.91    & 66.03  & -1.59                                \\ \hline
\multicolumn{1}{c|}{$\checkmark$}    & \multicolumn{1}{c|}{$\checkmark$}        & \multicolumn{1}{c|}{$\checkmark$}           & \multicolumn{1}{c|}{}          &             & 87.36    & 68.22  & 3.05                                 \\ \hline
\multicolumn{1}{c|}{$\checkmark$}    & \multicolumn{1}{c|}{}         & \multicolumn{1}{c|}{}            & \multicolumn{1}{c|}{$\checkmark$}         &             & 85.79    & 67.88  & 1.14                                \\ \hline
\multicolumn{1}{c|}{$\checkmark$}    & \multicolumn{1}{c|}{}         & \multicolumn{1}{c|}{}            & \multicolumn{1}{c|}{$\checkmark$}         & $\checkmark$           & 87.30    & 68.28  & 3.05                                 \\ \hline
\multicolumn{1}{c|}{$\checkmark$}    & \multicolumn{1}{c|}{$\checkmark$}        & \multicolumn{1}{c|}{$\checkmark$}           & \multicolumn{1}{c|}{$\checkmark$}         & $\checkmark$          & \textbf{88.34}    & \textbf{69.21}  & 5.06                                 \\ \hline
\end{tabular}
}
\end{center}
\vspace{-0.6cm}
\end{table}

\subsection{Information Gain from Synthetic Data}

\begin{figure}[t!]
    \begin{center}
    \includegraphics[width=1.0\linewidth]{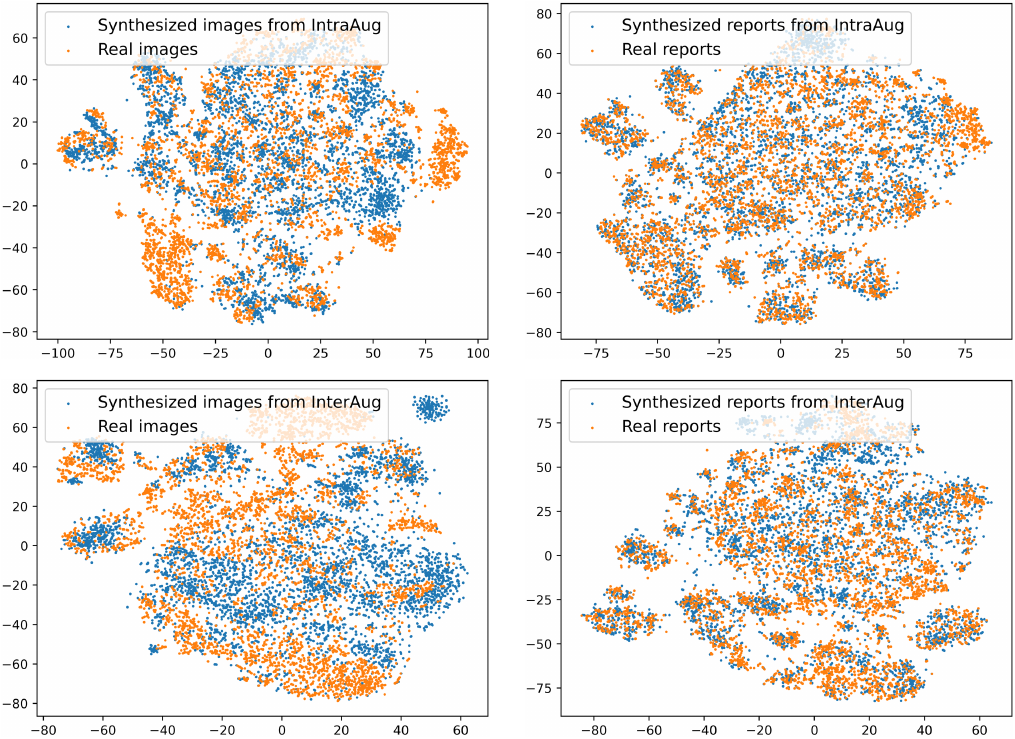}
    \end{center}
    \vspace{-0.6cm}
    \caption{T-SNE visualisation of image/report embeddings, comparing synthesised data from IntraAug and InterAug methods against real data from the MIMIC CXR dataset.}
    \label{fig:diversity}
    \vspace{-0.4cm}
\end{figure}

To illustrate the spread of the synthesised data, we randomly sample 5,000 image-report pairs from the augmented datasets created using the IntraAug and InterAug methods and the original MIMIC CXR dataset. We then extract the embeddings for both images and texts from these pairs. To visualise how the synthesised data compares to the original ones, we used t-SNE~\cite{van2008visualizing} to map the high-dimensional embeddings to a two-dimensional plane, as shown in Figure~\ref{fig:diversity}.
The t-SNE visualisation of synthesised and real data distributions suggests that our PairAug produce new, realistic variations that complement the existing MIMIC CXR dataset. The distributions are close in the embedding space yet with limited overlap, indicating an expansion of the dataset's diversity without diverging from authentic medical cases. This additional variety in the synthesised data could enhance model generalisation, as it introduces unique, realistic scenarios for robust deep learning training.

\subsection{Qualitative Results}
\begin{figure}[t!]
    \begin{center}
    \includegraphics[width=0.7\linewidth]{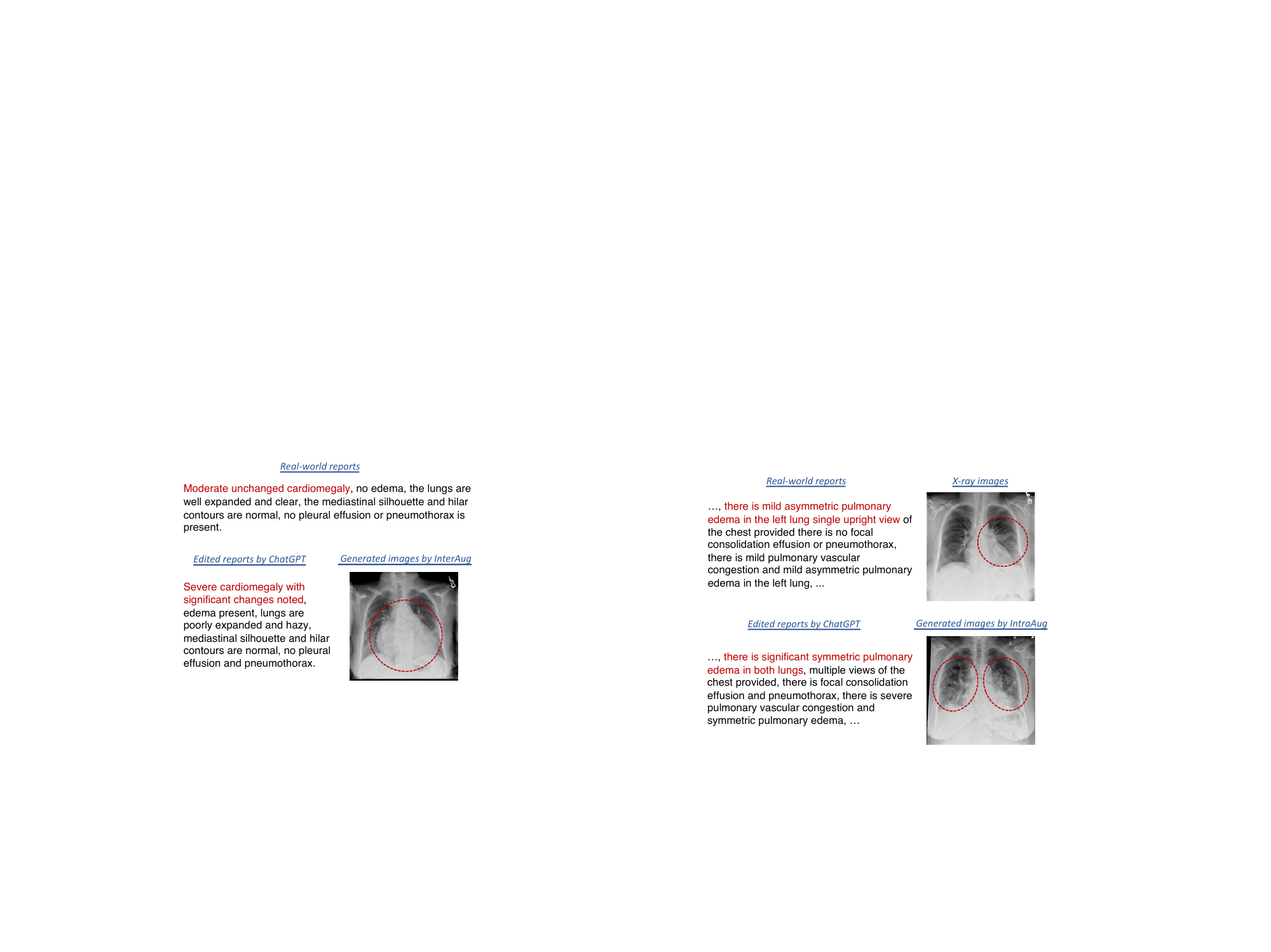}
    \end{center}
    \vspace{-0.6cm}
    \caption{Radiology report before and after editing by ChatGPT and the corresponding images generated by our InterAug. We highlight the specific areas in the radiology image with red bounding boxes and the corresponding descriptions in reports with the same colour.}
    \label{fig:vis_0}
    \vspace{-0.3cm}
\end{figure}

Figures~\ref{fig:vis_0} and~\ref{fig:vis_1} provide a visualisation of the results of our approach, revealing three interesting observations.
First, ChatGPT can effectively facilitate text augmentation. Given different prompts, ChatGPT can generate diverse reports and mimic the generation of novel clinical cases, thus gaining new information. However, note that ChatGPT occasionally outputs informal terms, such as ``\textit{homogeneous clearance}''. This highlights a potential area for future improvements, such as developing more refined prompts or filtering mechanisms to remove these casual terms. 
Secondly, our PairAug shows its capacity to generate medical images with a strong semantic alignment with the corresponding reports, as denoted by the red boxes in images and red words in reports.
Third, Figure~\ref{fig:vis_1} provide more evidence that our IntraAug can generate radiology images representing varying medical conditions for a specific patient. For instance, the manifestation of pulmonary edema in both lungs could potentially indicate a worsening condition compared to the initial stage, where the pulmonary edema is only present in the left lung. 
Some failure cases are shown in the supplementary.


%



\subsection{Human Radiologists Study}




We conducted a Visual Turing Test with medical experts, consisting of three distinct tasks to validate the authenticity of the image-text pairs generated by the PairAug method. 
Two radiology residents participated in this test. 
Task 1 involved 50 patient data sets, including real images and PairAug synthetic images. 
Task 2 involved 50 patient data sets, including real reports and PairAug synthetic reports. 
For these two Tasks, the experts were asked to identify each image/report as  ``real'' or ``synthetic/unsure''.
Task 3 involved 50 patient data sets, including synthetic pairs produced by PairAug. For this task, the experts were asked to assess the semantic alignment between the image and its corresponding report, categorising each pair as ``good'', ``poor'', or ``unsure''.

The results reflect the realism of medical image-report pairs generated by PairAug. With an average accuracy of 61\% and 50\% in Tasks 1 and 2, respectively, experts find distinguishing between real and synthetic images/reports challenging, suggesting a high degree of fidelity in the synthetic data. In addition, the average performance of 61\% in Task 3 suggests that although the coherence between modalities is generally convincing, there is room for further improvement.


\begin{figure}[t!]
    \begin{center}
    \includegraphics[width=0.75\linewidth]{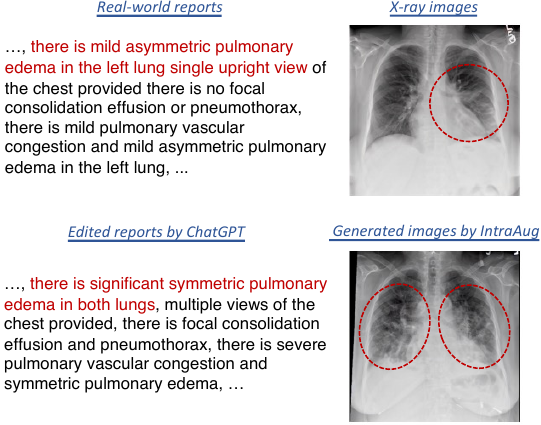}
    \end{center}
    \vspace{-0.6cm}
    \caption{Radiology image-report pair synthesised via IntraAug.}
    \label{fig:vis_1}
    \vspace{-0.5cm}
\end{figure}

\noindent \textbf{Limitation and Future Work}
The performance of PairAug relies heavily on the quality of the generation model employed. Here, we use the RoentGen model, pre-trained only on chest X-ray datasets, where the underlying training data is relatively biased and lacks diversity.
In future work, our goal is to train a model capable of generating radiology images from reports across various body parts, expanding our approach to a wider range of medical scenarios.



\section{Conclusion}

In this paper, we propose an approach called PairAug to address the challenge of acquiring paired image-text datasets in radiology.
PairAug consists of two branches: InterAug and IntraAug.
InterAug generates synthetic radiology images paired with plausible reports, creating new patient cases, while IntraAug focuses on generating diverse paired data for each individual.
We employ data pruning techniques to ensure high-quality data. Experimental results across various tasks show that PairAug outperforms baseline methods that only focus on either image or text expansion.

\noindent \textbf{Acknowledgements} Yutong Xie was supported by the Centre for Augmented Reasoning (CAR) project. Yong Xia was supported in part by the National Natural Science Foundation of China under Grants 62171377, and in part by the Ningbo Clinical Research Center for Medical Imaging under Grant 2021L003 (Open Project 2022LYKFZD06).


\renewcommand\thesection{\Alph{section}}
\renewcommand{\thetable}{\Alph{table}}
\renewcommand{\thefigure}{\Alph{figure}}
\setcounter{figure}{0}    
\setcounter{table}{0}    
\setcounter{section}{0}    

\section*{Appendix}

This document provides more discussions and experimental details to supplement the main submission. 
We organise the supplementary into the following sections. 
\begin{itemize}
\item In section~\ref{sec:details_exp_setup}, we depict more implementation details.
\item In section~\ref{sec:more_new_report}, we show more examples of newly generated reports by large language model.
\item In section~\ref{sec:failure_case}, we provide some visual results of failure cases. 
\item In section~\ref{sec:threshold}, we provide a discussion on the impact of hyper-parameters. 
\end{itemize}

\section{More Details of Experimental Setup}\label{sec:details_exp_setup}

\subsection{Computational Cost}

In our method, different parts have varying computational requirements, as indicated in Table~\ref{tab:tab_apex1}.
For the Pairwise Augmentation (PairAug) part, we use eight NVIDIA A100 GPUs, costing a total time of 24 hours. This high computational demand can be attributed to the inference complexity of diffusion models.
The medical visual-language pre-training (MedVLP) is trained on a less resource-intensive NVIDIA 3090 GPU, costing a total time of 5 hours.
The downstream tasks are also performed on an NVIDIA 3090 GPU, where the fine-tuning task takes no more than one hour for training and less than 1 second for testing.
Despite the high computational expenditure of the PairAug and MedVLP parts, the fast downstream training and online testing processes indicate the feasibility of incorporating our approach into regular clinical workflows.

\begin{table}[!b]
\small
\caption{Computational Cost.}
\vspace{-0.5cm}
\label{tab:tab_apex1}
\begin{center}
\renewcommand\arraystretch{1.0}
\setlength\tabcolsep{2.5pt}
\begin{tabular}{c|c|c|c}
\hline
& PairAug & MedVLP        & Downstream           \\ \hline
GPUs         & eight A100     & one 3090      & one 3090             \\ \hline
Time Cost   & 24 hours      & 5 hours      & one hour         \\ \hline
\end{tabular}
\end{center}
\end{table}

\begin{figure*}[t!]
    \begin{center}
    \includegraphics[width=0.95\linewidth]{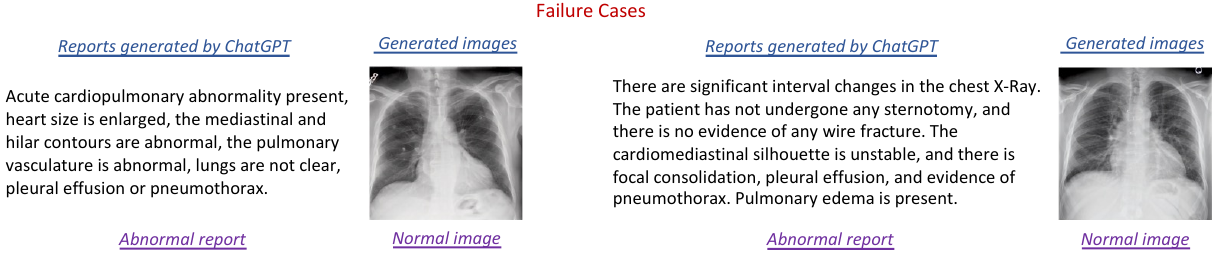}
    \end{center}
    \vspace{-0.5cm}
    \caption{Visual Results of Failure Cases}
    \label{fig:vis_2}
\end{figure*}

\begin{figure}[t!]
    \begin{center}
    \vspace{-4pt}
    \includegraphics[width=0.7\linewidth]{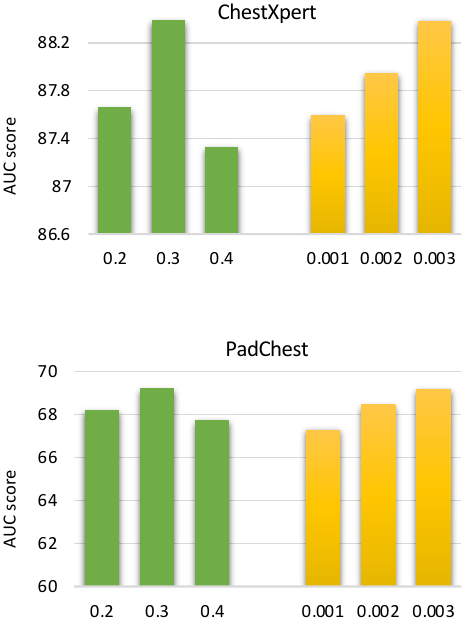}
    \end{center}
    \vspace{-13pt}
    \caption{AUC scores on ChestXpert and PadChest datasets vs. hyper-parameters values on thresholds $\tau$ (green bars) and $\epsilon$ (yellow bars).}
    \vspace{-10pt}
    \label{fig:hyper_parameter}
\end{figure}

\subsection{Experimental Details for Downstream}
We test the performance of learned pre-trained representations on three radiology-based downstream datasets:
(1) For CheXpert dataset~\cite{irvin2019chexpert}, we use its official test set for zero-shot evaluation aiming to classify each image into 5 five individual binary labels: atelectasis, cardiomegaly, consolidation, edema, and pleural effusion; 
(2) For PadChest dataset~\cite{bustos2020padchest} e adopt 39,053 chest X-rays annotated by board-certified radiologists for zero-shot evaluation. It has 193 disease image labels, including 174 radiographic findings and 19 differential diagnoses; and 
(3) For the RSNA Pneumonia dataset~\cite{shih2019augmenting}, we seek to classify each radiograph as negative or positive for pneumothorax. We divided the RSNA dataset into training, validation, and test sets with a ratio of 80\%/10\%/10\%, respectively. The input size is set to $224\times224$. We also employ online data argumentation to enlarge the training dataset. We optimise the downstream network with the AdamW~\cite{adamw} algorithm with cross-entropy loss and empirically set the initial learning rate to 0.0005, the batch size to 96, and the max epochs to 50.

For the zero-shot evaluation, we adopt a methodology inspired by the work~\cite{tiu2022expert}. This involves using labels from the test set to generate both positive and negative prompts for each condition. Specifically, for a given label, we create a positive prompt, such as ‘<label>’, and a corresponding negative prompt, ‘no <label>’, to facilitate the softmax evaluation process. The evaluation procedure is structured as follows: Initially, we calculate logits – a type of raw output from the last layer of neural networks before applying the softmax function – using both positive and negative prompts. For instance, for the label ‘atelectasis’, we compute logits for ‘atelectasis’ as the positive prompt and ‘no atelectasis’ as the negative prompt. Following this, we apply the softmax function to these logits, creating a probabilistic comparison between the positive and negative scenarios. Finally, the softmax probabilities derived from the positive logits are interpreted as the likelihood of the corresponding disease being present in the chest X-ray image.

\section{Samples of New Reports Generated by LLM}\label{sec:more_new_report}

We provide more samples of new radiology reports generated by ChatGPT. 
As shown in Figure~\ref{fig:vis_1}, given the prompt of ``\textit{Following is an original chest X-Ray report. Generate one possible augmentation that is limited to 50 words while conveying partial opposite meanings than the original report}'' with original chest X-Ray reports, ChatGPT exhibits the ability to produce varied reports. This ability to simulate the creation of new clinical scenarios contributes significantly to generating novel information continually.
Nevertheless, we must also acknowledge the limitations of the current report generation strategy. As highlighted in the primary manuscript, ChatGPT occasionally produces informal terminology. This observation underscores potential avenues for future enhancement, such as creating more sophisticated prompts or implementing filtering mechanisms to expunge these casual terms.
Moreover, the current strategy for generating reports relies on including original reports as inputs, which may constrain the diversity of the newly synthesised information. This limitation underscores the necessity to devise more diverse prompts or to develop a strategy that is not reliant on original reports, thereby enhancing the variety and richness of the augmented data.

\begin{figure*}[t!]
    \begin{center}
    \includegraphics[width=0.85\linewidth]{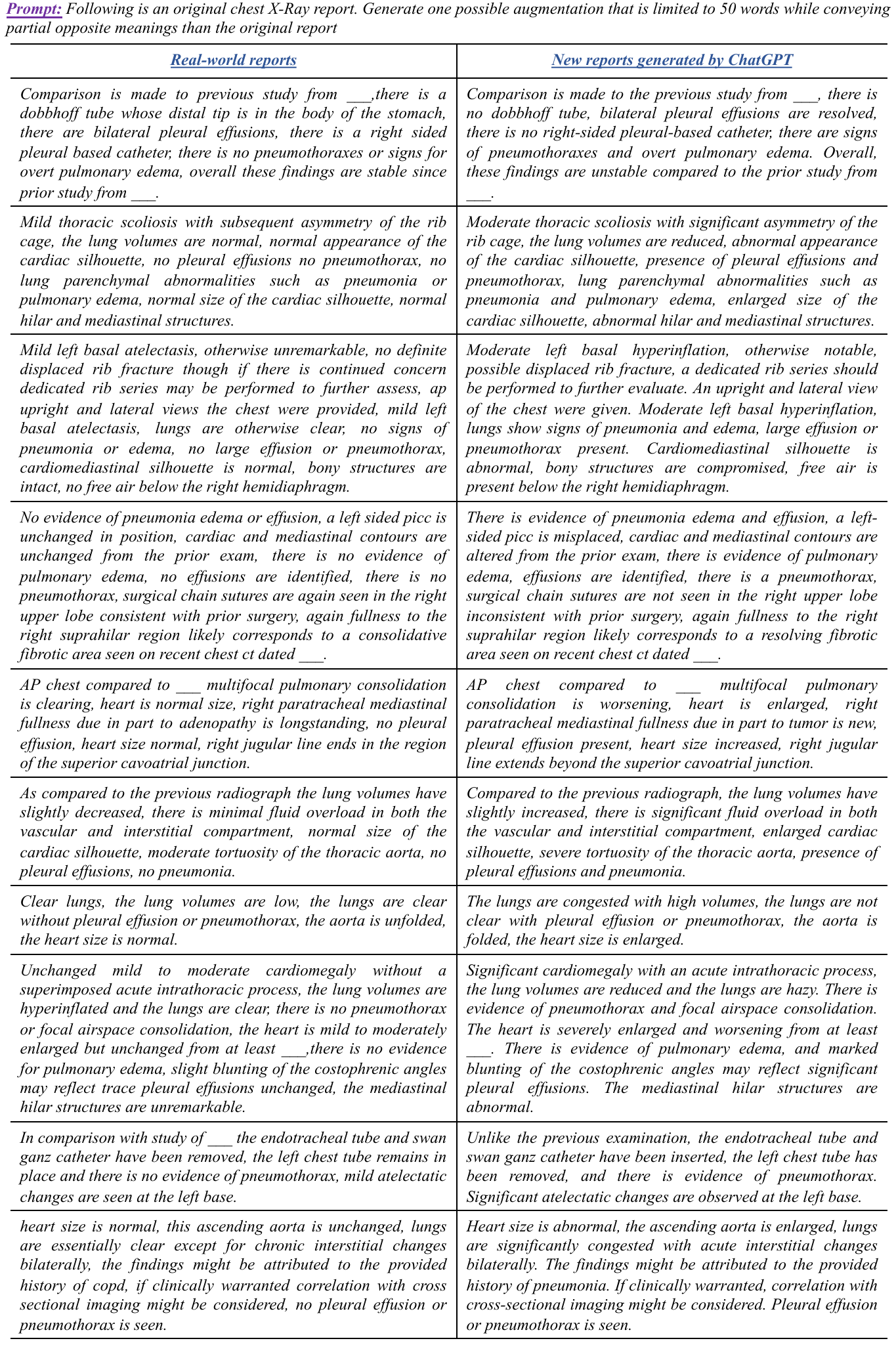}
    \end{center}
    \vspace{-0.6cm}
    \caption{More samples of new reports generated by ChatGPT. `\underline{\hbox to 4mm{}}' means anonymous processing.}
    \label{fig:vis_1}
\end{figure*}

\section{Some Visual Results of Failure Cases}\label{sec:failure_case}
In our study, some failure cases exhibited a clear disconnect between the semantic content of the generated report and the corresponding generated image, as shown in Figure~\ref{fig:vis_2}. These instances of misalignment can potentially be attributed to the limitations of the employed generation model, specifically, the RoentGen model~\cite{chambon2022roentgen}.
The RoentGen model's training regimen primarily focused on the 'impression' sections of the radiology reports. While this approach allowed it to develop a general understanding of the diagnostic conclusions drawn from the images, it might not have adequately exposed the model to the intricate details contained within the 'findings' sections. These sections often contain detailed descriptions of specific abnormalities present in the images, and the omission of this data in training might limit the model's ability to generate fully accurate and semantically aligned text and images.
Furthermore, the model's performance may also be affected by the imbalance in the representation of normal and abnormal findings in the training data. Particularly, the model might struggle to accurately represent images with multiple or rare abnormalities due to insufficient exposure during training.
These limitations also inspire further refinements in the generation model, including more detailed findings in the training stage, which might help improve the semantic alignment between the generated reports and images.


\section{Impact of Thresholds $\tau$ and $\epsilon$}\label{sec:threshold}
In Figure~\ref{fig:hyper_parameter}, we investigate the influence of the thresholds $\tau$ and $\epsilon$ in shaping the model's performance on the downstream tasks. 
Analysing the impact of $\tau$, we notice an increase in performance when we move the threshold from 0.2 to 0.4 on both the ChestXpert and PadChest datasets. This suggests that a higher threshold for $\tau$ allows us to filter out low-quality image-text pairs generated by the InterAug branch, thus enhancing the overall quality of the training data. However, further increasing $\tau$ to 0.4 results in a slight performance drop. This indicates that when the threshold becomes too high, the data pruning might become too aggressive, excluding potentially beneficial image-text pairs and causing a slight performance decrease.
Turning our attention to $\epsilon$, we observe a substantial performance increase on both datasets when the threshold increases from 0.001 to 0.003. This improvement suggests that a non-zero threshold for $\epsilon$ can help selectively incorporate the higher-quality synthesised data from the IntraAug branch. 
These results illustrate the importance of carefully setting the thresholds $\tau$ and $\epsilon$ for the model to use the synthesised data effectively. The optimal threshold settings would strike a balance: they would allow for the inclusion of beneficial synthesised data while filtering out potentially detrimental low-quality pairs, thus leading to improved performance on downstream tasks.


{
    \small
    \bibliographystyle{ieeenat_fullname}
    \bibliography{main}
}


\end{document}